\pdfoutput=1

\documentclass[11pt]{article}

\usepackage{acl}

\usepackage{times}
\usepackage{latexsym}
\usepackage{dirtytalk}
\usepackage{amsmath}
\usepackage{csquotes}
\usepackage{graphicx}
\usepackage{multirow}
\usepackage{makecell}
\usepackage{booktabs}
\usepackage{pgfplots}
\usepackage{enumitem}
\usepackage{hyperref}
\usepackage{amsfonts}
\usepackage{tabularx}
\usepackage{longtable}
\usepackage{subcaption}

\usepackage[T1]{fontenc}

\usepackage[utf8]{inputenc}

\usepackage{microtype}

\title{Measuring and Improving Semantic Diversity of Dialogue Generation}

\author{Seungju Han\;\;\;\; Beomsu Kim\;\;\;\; \textbf{Buru Chang}\thanks{\; Corresponding author} \\
  Hyperconnect \\
  \texttt{wade3han@snu.ac.kr},
  \{\texttt{beomsu.kim,buru.chang\}@hpcnt.com} \\
}

\begin{document}
\maketitle
\begin{abstract}
Response diversity has become an important criterion for evaluating the quality of open-domain dialogue generation models.
However, current evaluation metrics for response diversity often fail to capture the semantic diversity of generated responses, as they mainly consider lexical aspects of the generated responses.
In this paper, we introduce a new automatic evaluation metric to measure the semantic diversity of generated responses.
Through human evaluation, we demonstrate that our proposed metric captures human judgments on response diversity better than existing lexical-level diversity metrics.
Furthermore, motivated by analyzing an existing dialogue dataset, we propose a simple yet effective learning method that improves the semantic diversity of generated responses.
Our learning method weights training samples based on the semantic distribution of the training set.
We show that our learning method improves response diversity and coherency better than other baseline methods through automatic and human evaluation.
\end{abstract}
\section{Introduction}\label{sec:1_introduction}

Open-domain dialogue generation~\cite{sordoni2015neural,DBLP:conf/iclr/BordesBW17} has greatly progressed with the development of large-scale pretrained language models~\cite{radford2019language,roller2021recipes} in the last decade.
However, although dialogue generation models can produce fluent responses, they are also known for frequently generating dull and uninformative generic responses (e.g., "\textit{I don't know}"), degrading their engagingness~\cite{serban2016building,li2016diversity}.
To alleviate this problem, many studies~\cite{zhao2017learning,li2017adversarial,zhang2018generating} have been conducted to enhance the \textit{diversity} of generated responses, and response diversity has become an important criterion for evaluating the quality of generated responses.

\begin{figure}[t]
\centering
\includegraphics[width=0.95\columnwidth]{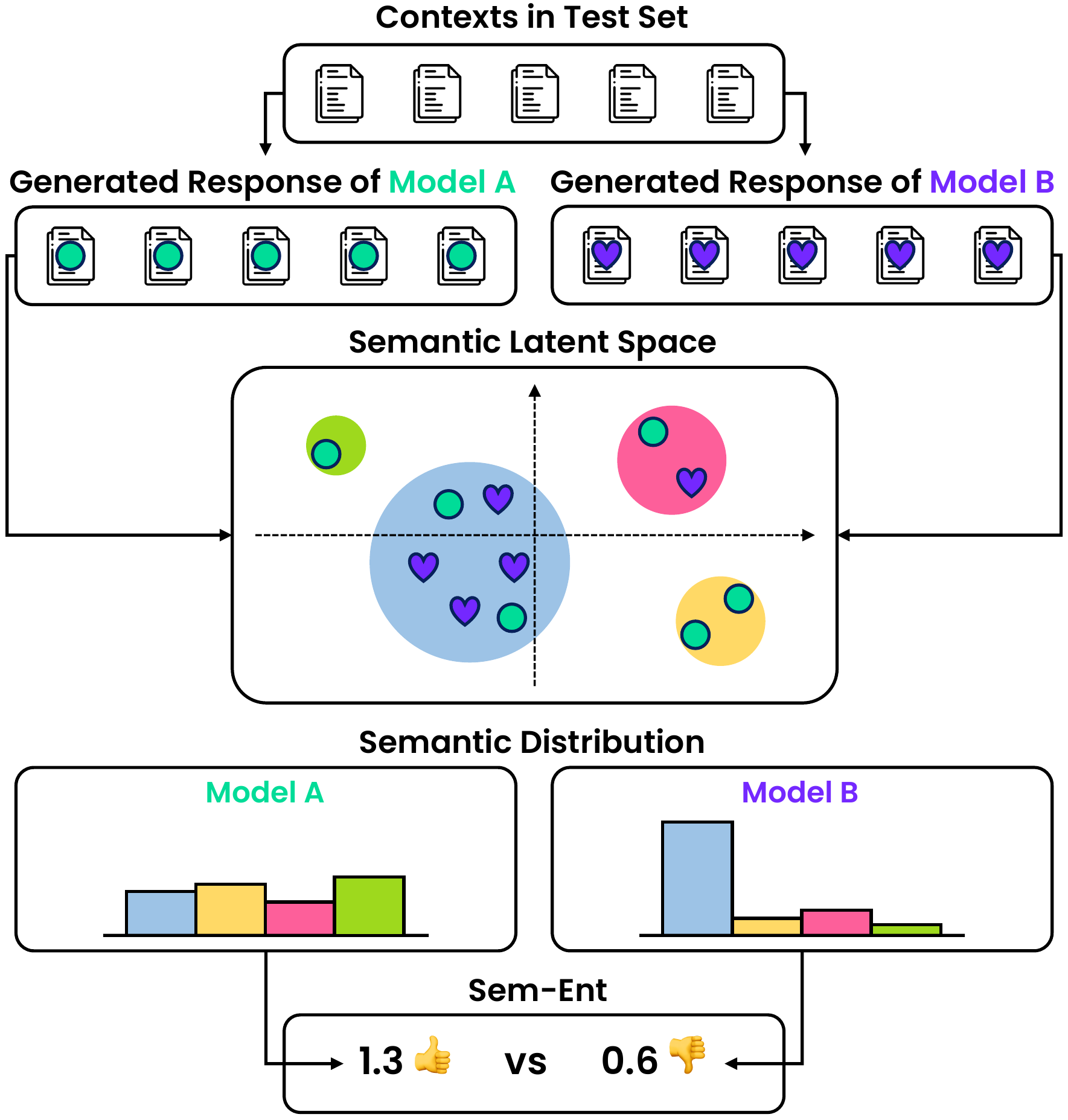}
\caption{
An illustration of our proposed Sem-Ent that measures semantic diversity based on the semantic distribution of generated responses.}
\vspace*{-1.5em}
\label{fig:1_introduction}
\end{figure}
The current evaluation protocol has relied on lexical-level evaluation metrics such as Distinct-$n$ (Dist-$n$)~\cite{li2016diversity} and Entropy-$n$ (Ent-$n$)~\cite{serban2017hierarchical} to measure the diversity of generated responses.
However, according to recent studies~\cite{tevet2021evaluating,stasaski2022semantic}, these lexical-level evaluation metrics often fail to capture \textit{semantic diversity} since responses including similar words can have different semantics and responses with different words can have similar semantics~\cite{yarats2018hierarchical}.

In this paper, we propose \textit{\textbf{Sem-Ent}} (\underline{\textbf{Sem}}antic-\underline{\textbf{Ent}}ropy), which is a new automatic evaluation metric for measuring the semantic diversity of generated responses.
Sem-Ent first maps generated responses into a semantic latent space using a pretrained language model (e.g., DialoGPT~\cite{zhang2020dialogpt} and BERT~\cite{devlin2019bert}). 
Then, the evaluation metric measures the semantic diversity of generated responses by calculating how the responses are evenly distributed in the semantic latent space based on entropy, as shown in Figure~\ref{fig:1_introduction}.
Through human evaluation, we demonstrate that Sem-Ent is more highly correlated with human judgments on response diversity than existing lexical-level evaluation metrics.

Furthermore, we propose a simple yet effective learning method of dialogue generation models to improve semantic diversity of generated responses.
We observe that the semantic distribution of responses in a dialogue dataset is highly imbalanced, leading dialogue generation models to produce semantically less diverse responses.
To address this problem, our proposed method, \textbf{\textit{DRESS}} (\underline{\textbf{D}}iversifying \underline{\textbf{RES}}ponses \underline{\textbf{S}}emantically), learns more about responses with rare semantics and learn less about responses with frequent semantics.
From this, dialogue generation models could produce more semantically diverse responses.
Experiments on two benchmark datasets demonstrate that DRESS shows better semantic diversity compared to state-of-the-art baseline methods, along with the gain in response coherency.
Interestingly, DRESS also achieves better performance in lexical-level diversity metrics than baselines, even though it focuses only on improving the semantic diversity.
Moreover, human evaluation results show the effectiveness of DRESS, where DRESS outperforms all baseline methods in appropriateness and informativeness of generated responses.

\noindent
\textbf{Our Contributions:}
(1) A new automatic evaluation metric for measuring semantic diversity (Sem-Ent), which is highly correlated with human judgments on response diversity.
(2) A simple yet effective learning method of dialogue generation models (DRESS) for improving the semantic diversity of generated responses.
(3) Experiments on two benchmark datasets, showing that DRESS outperforms the baseline methods in both semantic diversity and lexical-level diversity.
(4) An implementation of Sem-Ent will be released, contributing to the community of open-domain dialogue generation.
\section{Related Work}\label{sec:2_related_work}

\subsection{Enhancing Response Diversity}\label{subsec:2_1_enhancing_response_diversity}
Since generating dull and uninformative responses is a well-known and important problem in open-domain dialogue \citep{vinyals2015neural, li2016diversity}, numerous methods have been proposed to address this issue.
The maximum mutual information objective function is utilized to penalize generic responses and improve the diversity of generated responses~\cite{li2016diversity,li2016deep,zhang2018generating,zhang2020dialogpt}.
Another line of work improves diversity by modeling the one-to-many relationship of open-domain dialogue using latent variables to generate multiple and diverse responses~\citep{serban2017hierarchical,zhao2017learning,bao2020plato,bao2020plato2, ijcai2019-683,zhang2019improve,gao2019jointly}.
Some methods selectively penalize frequent responses by removing them from the training set \citep{csaky2019improving} or applying negative training to frequent responses \citep{he2020negative}.
Using different decoding algorithms can improve the response diversity; \citet{li2016simple} and \citet{vijayakumar2018diverse} directly modify the beam search algorithm to promote the response diversity.
Sampling-based decoding algorithms such as top-$k$ sampling \citep{fan2018hierarchical} and nucleus sampling \citep{holtzman2019curious} are also known to improve the diversity of generated responses.
\citet{wang2021diversifying} diversify responses by adaptively modifying the target token distribution with a lightweight decoder to prevent the model from being over-confident.

\subsection{Metrics for Capturing Response Diversity}\label{subsec:2_2_diversity_metrics}
Response diversity metrics for open-domain dialogue generation models can mainly be categorized into two groups.
Referenced metrics~\cite{zhao2017learning, gao2019jointly} use the reference responses provided by human annotators to capture the response diversity by computing a recall value based on various similarity metrics such as BLEU and embedding similarity.
On the other hand, unreferenced metrics measure the response diversity without using reference responses generated by human annotators.
Unreferenced metrics are more widely adopted than referenced metrics because they can measure response diversity even in the absence of reference responses.
Dist-$n$ \citep{li2016diversity} measures the response diversity with the fraction of distinct $n$-grams over possible $n$-grams in all generated responses.
Ent-$n$ metric \citep{serban2017hierarchical,zhang2018generating} is suggested to improve the Dist-$n$ metric by taking the frequency difference of $n$-grams into account.
Low-Frequency (LF)~\citep{li2019data} calculates the frequency of low-frequency words in generated responses as the response diversity.

\noindent
\textbf{Semantic diversity.}
Recently, several studies have focused on the semantic diversity of generated responses.
\citet{tevet2021evaluating} release the McDiv benchmark to evaluate the semantic diversity metrics and \citet{stasaski2022semantic} propose a new semantic diversity metric, natural language inference (NLI) diversity, leveraging pretrained NLI models~\cite{bowman2015large}.

The major difference between our Sem-Ent and NLI diversity is that NLI diversity can only capture the semantic diversity of generated responses for a single context, while Sem-Ent measures the overall semantic diversity of generated responses for multiple contexts of the test set.
This is an important distinction since the latter provides insight into how well generated responses vary depending on which context is provided as an input while the former cannot.
To see the shortcoming of NLI diversity more clearly, take the following an example: suppose that given a context $c_{a}$ as an input, a dialogue generation model generated responses $\{r_{a,1},r_{a,2},\cdots\}$ that are "semantically diverse" according to NLI diversity.
Now, further suppose that given another context $c_{b}$, the model generates responses $\{r_{b,1},r_{b,2},\cdots\}$ that are also "semantically diverse" among themselves but appear similar to the responses $\{r_{a,1},r_{a,2},\cdots\}$ produced for the context $c_{a}$.
In such a case, NLI diversity cannot capture the fact that the generated responses $\{r_{a,1},\cdots\}$ and $\{r_{b,1},\cdots\}$ for the contexts $c_{a}$ and $c_{b}$ are semantically similar despite the two contexts being different contexts; Sem-Ent can because it measures the semantic diversity of generated responses for a set of different contexts of the test set.  
In the next Section~\ref{sec:3_metric}, we will describe our proposed semantic diversity metric, Sem-Ent, in detail.

\section{Measuring Semantic Diversity}\label{sec:3_metric}
\subsection{Sem-Ent}\label{subsec:3_1_sem_ent}
Let $\mathcal{D}$ = $\{(c_i,r_i)\}_{i=1}^{m}$ denote a training set consisting of $m$ dialogues where $c_i$ and $r_i$ denote the context and its response of the $i$-th dialogue, respectively.
Dialogue generation is to generate a response $r$ for a given context $c$.

We are motivated by recent empirical observations that responses can be clustered by the semantic similarity between the responses~\cite{ko2020generating,gao2020dialogue}.
By following \citet{csaky2019improving, pillutla2021mauve}, we cluster responses in $\mathcal{D}$ by utilizing a pretrained language model.
Here, we select DialoGPT~\cite{zhang2020dialogpt} as the language model.
Each response $r_i \in \mathcal{D}$ is turned into a semantic representation $e(r_i)$ by the language model, and then $k$ semantic clusters are formed from the semantic representations by the $k$-means algorithm~\cite{lloyd1982least}.
Let $\mathcal{C}$ denote a set of the obtained $k$ semantic clusters.

Consider a test set $\Tilde{\mathcal{D}}$ = $\{(\Tilde{c}_i,\Tilde{r}_i)\}_{i=1}^{n}$ consisting of $n$ dialogues. 
During evaluation, a dialogue generation model $M$ generates responses $\mathcal{R}^M$ = $\{r^M_i\}_{i=1}^{n}$ for the contexts $\{\Tilde{c}_i\}_{i=1}^{n} \in \Tilde{\mathcal{D}}$, respectively.
To compute semantic diversity, Sem-Ent requires a semantic distribution $P(\mathcal{R}^M)$, but there is no direct way to obtain the exact distribution.
Thus, we approximate the semantic distribution $P(\mathcal{R}^M)$ as $\Tilde{P}(\mathcal{R}^M)$ = $\big[\Tilde{p}(1);\cdots;\Tilde{p}(k)\big]$ using the the semantic clusters $\mathcal{C}$ as follows:
\begin{align}
    \Tilde{p}(j)=\frac{1}{n} \sum_{i=1}^n \mathbb{I}\big(\phi_{\mathcal{C}}(e(r^M_i)) = j \big),
\end{align}\label{eq:cluster}
where $\phi_{\mathcal{C}}(r) \in \{1,\cdots,k\}$ is a cluster mapping function that returns the cluster index of a response $r$ from $\mathcal{C}$.
$\Tilde{p}(j)$ is the probability of the $j$-th cluster, indicating how many generated responses are assigned to the $j$-th semantic cluster.

Sem-Ent is an entropy of $\tilde{P}(\mathcal{R}^M)$ that approximates the semantic distribution $P(\mathcal{R}^M)$ as follows:
\begin{align}
    \text{Sem-Ent}(\mathcal{R}^M) = - \sum_{j=1}^k \tilde{p}(j) \cdot \log\tilde{p}(j).
\end{align}\label{eq:sem_ent}
Interpretation of Sem-Ent is quite straightforward: Sem-Ent gets lower when the semantic distribution gets more imbalanced, i.e., when models generate responses belonging to only several specific semantic clusters.
Conversely, Sem-Ent gets the highest value of $\log k$ when generated responses are uniformly distributed to each semantic cluster.

\subsection{Correlation with Human judgment}\label{subsec:3_2_correlation_with_human_judgment}
\begin{table*}[t]
\centering
\footnotesize
\begin{tabular}{cclllll}
\toprule
 \textbf{Metric} & \textbf{Correlation} & \textbf{Dist-3} & \textbf{Ent-3} & \textbf{LF} & \textbf{MAUVE} & \textbf{Sem-Ent} \\
\midrule 
\multirow{2}{*}{Diversity/BT} &
Pearson & 0.348 (0.399) & 0.702 (0.052) & -0.232 (0.580) & 0.134 (0.750) & \textbf{0.810 (0.015)} \\
& Spearman & 0.381 (0.352) & 0.667 (0.071) & 0.000 (1.000) & 0.547 (0.160) & \textbf{0.762 (0.028)}  \\
\midrule
\multirow{2}{*}{Interesting/BT} &
Pearson & 0.261 (0.533) & 0.671 (0.068) & -0.260 (0.533) &  0.098 (0.817) & \textbf{0.789 (0.020)}\\
& Spearman & 0.381 (0.352) & \textbf{0.714 (0.047)} & 0.048 (0.911) & 0.523 (0.182) & 0.667 (0.020) \\
\bottomrule
\end{tabular}
\caption{Correlation of diversity matrices with human judgments on response diversity.
BT denotes the Bradley-Terry score for a pairwise
human evaluation and the value inside the parenthesis indicates p-value.
We set the number of semantic clusters as $k$=20.
Evaluation results with different $n$ for Dist-$n$ and Ent-$n$ are reported in Appendix~\ref{subsec:a_4_further_results}.
}
\vspace{-1em}
\label{tab:1_sem_ent_human_judgment}
\end{table*}

We conduct a human evaluation to demonstrate that Sem-Ent successfully captures human judgments on response diversity.

\noindent\textbf{Experimental Setup.}
We borrow a pairwise experimental setup of \citet{pillutla2021mauve} for analyzing the correlation between diversity metrics and human judgments.
Our evaluation is based on the observation that the degree of response diversity varies depending on the types of generative models and decoding algorithms~\cite{holtzman2019curious,tevet2021evaluating}.
From this, we first prepare eight different response generation settings from two generation models (Blender-90M~\cite{roller2021recipes} and BART-large~\cite{lewis2020bart}) and four decoding algorithms (greedy, beam, top-k sampling, and nucleus sampling).
We then obtain 28 pairs of generation settings from the eight response generation settings.

For each pair, we randomly choose ten contexts from the test set of a DailyDialog dataset~\cite{li2017dailydialog} and generate two response sets using the two generation settings, respectively, for the ten contexts.
Human annotators are asked to select which response set is better in two criteria, diversity and interestingness, using a 5-point Likert scale.
We obtain 25 pairwise annotations for each pair of response generation settings.
These annotation results are converted into each response generation setting's score by using the Bradley-Terry model \cite{marden1996analyzing}.
By Bradley-Terry model, the probability of the outcome $i \succ j$ is calculated as $p(i \succ j) = e^{\theta_i} / (e^{\theta_i} + e^{\theta_j})$ when parameters $\theta_1, \cdot, \theta_n$, for two items $i$ and $j$ are given.
For more details about the Bradley-Terry model, please refer to choix manual\footnote{https://github.com/lucasmaystre/choix}.

We measure the correlation between the Bradley-Terry score and diversity metrics to check how each metric correlates with the human judgments on each criterion.
More details about human evaluation are included in Appendix~\ref{sec:a_details_human_evaluation_sement}.
\newline
\textbf{Baseline Metrics.}
We compare Sem-Ent with existing lexical-level response diversity metrics: Dist-$n$ \citep{li2016diversity}, Ent-$n$ \citep{serban2017hierarchical, zhang2018generating} and LF \citep{li2019data}.
We also include the recently proposed MAUVE~\citep{pillutla2021mauve} as a baseline metric.
MAUVE shares some properties with Sem-Ent such that it evaluates the distributional property of generated responses with semantic latent representations.
However, it is designed to measure the divergence of generated responses from human responses, not directly measuring response diversity.
We compare Sem-Ent to MAUVE to verify that our Sem-Ent is more suitable for measuring the response diversity in open-domain dialogue generation.
Note that we do not set NLI diversity as a baseline because it is incompatible with our human evaluation which measures the overall semantic diversity.
\newline
\textbf{Results.}
Table \ref{tab:1_sem_ent_human_judgment} shows the correlation between the human judgments and diversity metrics in terms of Pearson and Spearman rank correlation.
Our Sem-Ent shows the highest Pearson and Spearman rank correlation with human judgments on response diversity compared to other evaluation metrics with a significant margin.
Especially, Dist-$n$, the most commonly used metric for response diversity, shows a much lower correlation (0.348) compared to Sem-Ent (0.810).
These results support that Sem-Ent is a good surrogate for estimating human judgments on response diversity and strongly suggest that analyzing the semantic diversity of generated responses is crucial for capturing human perception of response diversity.
Moreover, MAUVE shows a lower correlation with human judgments on response diversity.
This result implies that generated responses that have similar representations to human responses (i.e., high MAUVE scores) are not always semantically diverse since human responses are also often generic~\cite{csaky2019improving} (further analyzed in Section~\ref{subsec:4_1_diagnose}).

We also observe that Sem-Ent shows a high correlation with human judgment on interestingness~\cite{see2019makes}; 
Ask annotators about how interesting or boring did they find about the generated response.
Sem-Ent has a similar correlation to Ent-$n$ and shows a substantially higher correlation than Dist-$n$, LF, and MAUVE.
We believe that semantically diverse responses could improve the interestingness of dialogue generation models and Sem-Ent could somewhat capture human judgments on this response interestingness.

\noindent\textbf{Robustness of Sem-Ent to the Choice of Configuration.}
Sem-Ent could be affected by the changes in the configurations used for calculating the score, such as the types of language models and the number of semantic clusters $k$.
We conduct an additional evaluation to examine the robustness of Sem-Ent to the choice of the configurations,
By changing the configurations, we obtain multiple Sem-Ent scores of generated responses and then analyze the correlation between the scores.
The more details of the additional evaluation and results are described in Appendix~\ref{sec:b_robustness}.
We observe that Sem-Ent shows a high correlation among the obtained scores, verifying the robustness of Sem-Ent against the choice of configurations.

\section{Improving Semantic Diversity}\label{sec:4_method}
\begin{figure}[t]
\centering
\includegraphics[width=\columnwidth]{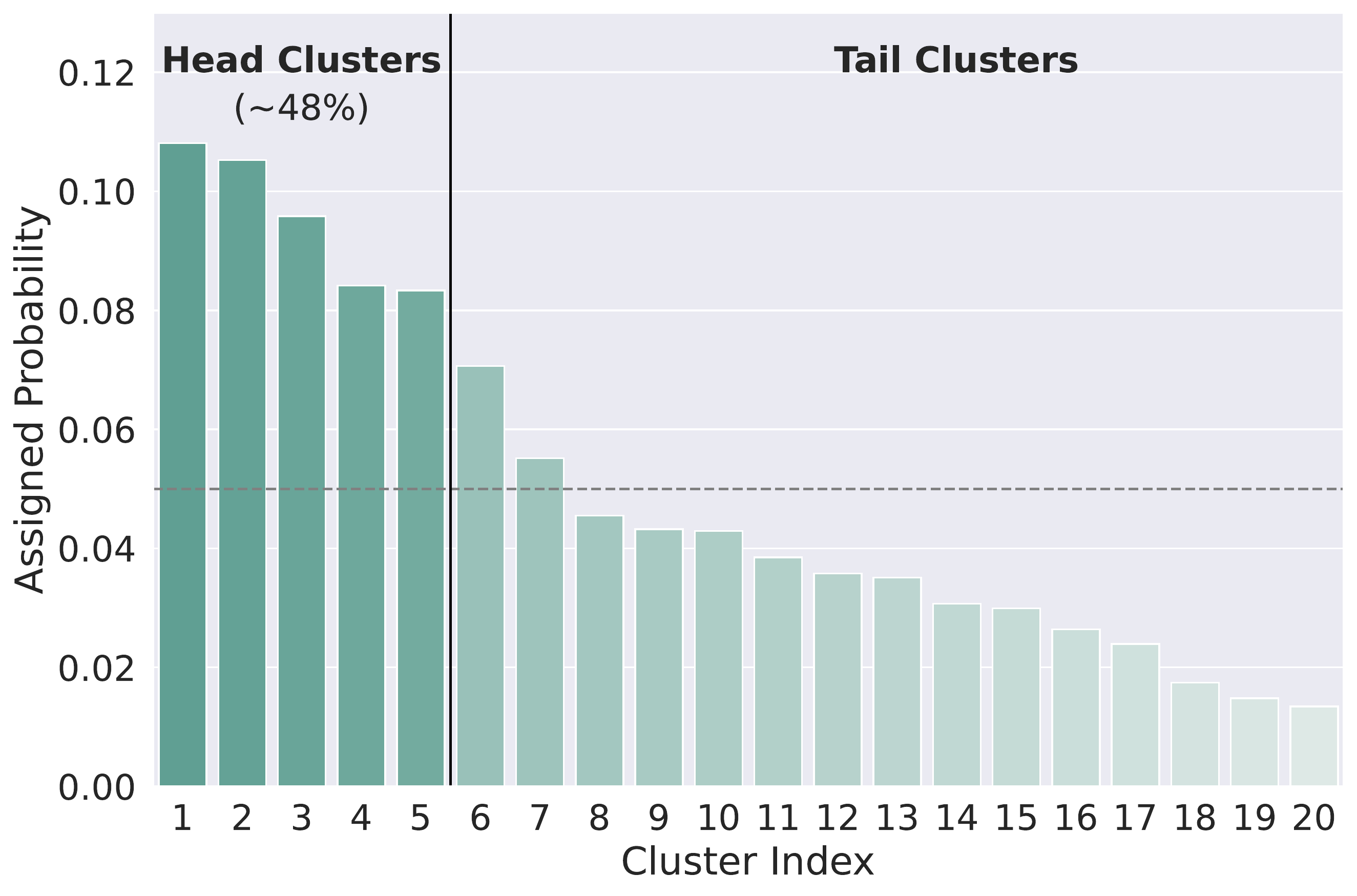}
\caption{Semantic distribution of responses in the training set. Semantic clusters are sorted in the descending order of the assigned probabilities. The dashed line indicates the uniformly distributed probability, 0.05.}
\label{fig:2_dataset}
\end{figure}
\begin{table}[t]
\centering
\footnotesize
\begin{tabular}{c|p{0.78\columnwidth}}
\toprule
\textbf{Index} & \textbf{Responses} \\
\midrule 
\multirow{4}{*}{1} & 
 $\boldsymbol{\cdot}$ Tell me . \\
& $\boldsymbol{\cdot}$ You're welcome . \\
& $\boldsymbol{\cdot}$ Bye . \\
& $\boldsymbol{\cdot}$ What's up ? \\
\midrule

\multirow{4}{*}{2} & 
 $\boldsymbol{\cdot}$ Yeah . I know . \\
& $\boldsymbol{\cdot}$ Thank you . \\
& $\boldsymbol{\cdot}$ That's cool . \\
& $\boldsymbol{\cdot}$ Not yet . \\
\midrule
\multirow{7}{*}{19} &
$\boldsymbol{\cdot}$ Yes . Will you also make copies and file them using both methods ? \\
& $\boldsymbol{\cdot}$ you should probably call the IT department and have them check your computer for virus . \\
& $\boldsymbol{\cdot}$ I see . Well , can I have a look at your phone ? Unfortunately , this phone can ' t be used in the US . it ' s not compatible with our 3G network . \\
\midrule
\multirow{7}{*}{20} &
$\boldsymbol{\cdot}$ A driver ’s license or something showing that you live in this city .\\
& $\boldsymbol{\cdot}$ I want to change a new car . I like Honda best , especially the red one . But it is too expensive . \\
& $\boldsymbol{\cdot}$  We use a vacuum cleaner that removes all the dirt , and we throw away all of the trash that we can
find . \\
\bottomrule
\end{tabular}
\caption{Response examples of the semantic clusters. \textit{Index} column indicates the index of semantic cluster in Figure \ref{fig:2_dataset}.}
\vspace*{-1em}
\label{tab:2_head_and_tail_examples}
\end{table}
\subsection{Diagnosing the Semantic Distribution of Dialogue Dataset}\label{subsec:4_1_diagnose}
As described in Section~\ref{sec:3_metric}, semantic distribution of responses provides a crucial clue for understanding the diversity of the responses.
Therefore, we analyze the semantic distribution of the responses $\mathcal{R}$ in the training set of the DailyDialog dataset.
Figure~\ref{fig:2_dataset} shows that the semantic distribution of $\mathcal{R}$ is highly skewed -- almost half of the responses fall into the top five frequent clusters (head clusters).
Moreover, the frequent clusters tend to contain more generic and dull responses compared to infrequent clusters (tail clusters), as shown in Table \ref{tab:2_head_and_tail_examples}.
Conversely, responses in the infrequent clusters have a wider variety of topics, intents, and diverse vocabularies.
Since the training set is skewed towards semantically generic and dull responses, naively training with this data will lead to the low semantic diversity of generated responses.

\subsection{DRESS}
We introduce a simple yet effective learning method of dialogue generation models for improving semantic diversity, DRESS, which addresses the problem of the imbalanced semantic distribution by weighting the responses in the training set.
The purpose of DRESS is simple: inducing generation models to learn more about responses in the infrequent semantic clusters and learn less about responses in the frequent semantic clusters.
To this end, DRESS modifies the learning objective into the weighted loss function and applies Negative Training~\cite{he2020negative, li2020don} to the modified objective.

A conventional dialogue generation model is trained by optimizing an NLL (negative log-likelihood) objective as follows:
\begin{align}
    L_{NLL}(D) = -\sum_{i=1}^m \log p_\theta(r_i|c_i),
\end{align}
where $\theta$ indicates parameters of dialogue generation models.
Instead of using vanilla NLL objective, we propose to utilize weighted NLL objective in DRESS using weight of responses $w(r_i)$:
\begin{align}
    L_{DRESS}(D) = -\sum_{i=1}^m w(r_i)\cdot\log p_\theta(r_i|c_i).
\label{eq:4_weighted_nll}
\end{align}
The goal of weighted NLL objective is to assign smaller weights to the responses in frequent semantic clusters and assign bigger weights to responses in infrequent semantic clusters to balance the semantic distribution.
To meet this condition, the weighting function $w(r)$ should satisfy the constraint: if $\tilde{p}(\phi_c(e(r_i))) \leq \tilde{p}(\phi_c(e(r_j)))$, then $w(r_i) \geq w(r_j)$.
Inspired by focal loss~\cite{lin2017focal} which is used in the long-tail classification problem~\cite{liu2019large,hong2021disentangling}, we calculate $w$ as follows:
\begin{align}
    w(r) = \big(1 - \tilde{p}(\phi_c(e(r)))\big)^\gamma,
\end{align}\label{eq:weighting}
where $\gamma$ is a hyperparameter for controlling a degree of weighting (higher $\gamma$ means more intense weighting).

Moreover, to penalize responses in frequent semantic clusters intensively, we jointly utilize Negative Training (NT)~\cite{he2020negative,li2020don} with the weighted objective function.
For every epoch, the model generates responses to each given context.
If generated responses are included in head clusters, then those generated responses are assumed as negative examples, i.e., assigning $w(r) = -1$.
\section{Experiments}\label{sec:5_experiments_dress}
\subsection{Experimental Setup}\label{subsec:5_1_experimental_setup}
We conduct experiments to demonstrate that the proposed DRESS successfully improves response diversity. 
\newline
\textbf{Dataset.}
We utilize two open-domain dialogue datasets: DailyDialog~\citep{li2017dailydialog} and OpenSubtitles~\citep{lison2016opensubtitles2016}.
DailyDialog consists of 13K dialogues which includes 87K context-response pairs, and we split the dialogues into train/valid/test sets in 8:1:1.
The test set of DailyDialog contains 6.7K context-response pairs.
OpenSubtitles is a large corpus containing movie scripts, and we use the version released in 2018 with 100K context-response pairs for the training and validation sets.
We get rid of context-response pairs whose response is shorter than five words from the original test set and randomly sample 10K pairs as test data.
\newline
\textbf{Automated Metrics.}
As the goal of diversity-promoting dialogue generation models is to generate diverse responses without hurting the coherency of responses, we focus on two criteria: response diversity and coherency.
For measuring response diversity, we use both lexical-level diversity metrics (Dist-$n$, Ent-$n$, and LF) and a semantic diversity metric (Sem-Ent).
For measuring response coherency, we employ MaUdE \citep{sinha2020learning}, an unreferenced dialogue response evaluation metric that shows a high correlation with human judgments on the fluency of generated responses.
\newline
\textbf{Human Evaluation.}
We further conduct a pairwise comparison through the human evaluation for evaluating the general conversation ability of generation models since automatic evaluations are sometimes not trustworthy.
We use Amazon Mechanical Turk to collect the annotations.
Following many previous studies~\citep{gao2019jointly,he2020negative,han-etal-2022-understanding}, each annotator evaluates which response is better in terms of \textit{Appropriateness} for measuring response coherency and \textit{Informativeness} for evaluating whether the given response has meaningful information relevant to its given context.
We collect annotations for 50 test cases per each model pair, and three annotators rate each test case to improve the robustness of the evaluation result.
Note that \textit{Diversity} cannot be assessed in this pair-wise evaluation setup.
More details about evaluation protocol (e.g., interface for collecting annotation) are described in Appendix~\ref{sec:c_details_human_evaluation_dress}.

\subsection{Baseline Methods}\label{subsec:5_2_baseline_methods}
\textbf{\textit{MMI}} \citep{li2016diversity} increases response diversity by maximizing the mutual information between context and response.
We utilize the MMI-antiLM as our MMI baseline.
\newline
\textbf{\textit{CVAE}} \citep{zhao2017learning} builds the response generation process as a conditional variational auto-encoder of a response with dialogue context to increase response diversity.
\newline
\textbf{\textit{EDF}} \citep{csaky2019improving}
enhances response diversity by filtering out context-response pairs that increase one-to-many or many-to-one problems in the training dataset.
We use source side entropy to filter the pairs.
\newline
\textbf{\textit{NT}}~\citep{he2020negative}  directly penalizes the generation of generic responses by applying reverse direction gradient for the losses of the generic responses, leading to maximizing the loss rather than minimizing it.
\newline
\textbf{\textit{AdaLabel}} \citep{wang2021diversifying} alleviates the over-confidence problem of generation models to improve response diversity by dynamically smoothing the target token distribution with an auxiliary lightweight decoder.

\begin{table*}[t]\setlength{\tabcolsep}{0.4em}
\centering
\footnotesize
\begin{tabular}{clccccccccc}
\toprule
Backbone& Method & Dist-1 & Dist-2 & Dist-3 & Ent-1 & Ent-2 & Ent-3 & LF & MaUdE & \textbf{Sem-Ent} \\
\midrule
\multirow{8}{*}{\shortstack[c]{Blender-90M\\(DailyDialog)}} & Vanilla &0.0453 & 0.2103 & 0.3881 & 7.1322 & 10.7502 & 12.3950 & 0.2234 & 0.8489 & 2.5486\\
& MMI &0.0349 & 0.1677 & 0.3069 & 7.0730 & 10.3806 & 11.9808 & 0.2155 & 0.8208 & 2.5784\\
& CVAE & 0.0471 &	\underline{0.2389} &	\underline{0.4459}& 	7.4074& 	11.2797&	12.9969&	0.2449& 0.8552	&	2.6261\\
& EDF & \underline{0.0473} & 0.2271 & 0.4226 & 7.2888 & 11.0283 & 12.7132 & 0.2402 & \underline{0.8593} & 2.5872\\
& NT & \textbf{0.0475} & 0.2351 & 0.4422 & 7.3994 & 11.2561 & 13.0111 & 0.2467 & \textbf{0.8597}& 2.6434\\
& AdaLabel & 0.0377 & 0.1982 & 0.3915 & 7.1546 & 10.8772 & 12.6829 & 0.2158 & 0.8443 & 2.6038\\
\cmidrule{2-11}
& DRESS(-NT) & 0.0445 & 0.2295 & 0.4360 & 7.4560 & \underline{11.3273} & \underline{13.1028} & \underline{0.2474} & 0.8460 & \underline{2.7576}\\
& \textbf{DRESS} & 0.0460 & \textbf{0.2404} & \textbf{0.4571} & \textbf{7.5468} & \textbf{11.5094} & \textbf{13.3060} & \textbf{0.2576} & 0.8575 & \textbf{2.7819}\\ 
\midrule
\multirow{8}{*}{\shortstack[c]{BART-large\\(DailyDialog)}} & Vanilla & 0.0462 & 0.2168 & 0.4056 & 7.3913 & 11.2075 & 12.8648 & 0.2593 & 0.8854 & 2.4251\\
& MMI & 0.0497 & 0.2329 & 0.4355 & 7.4748 & 11.4060 & 13.0898 & 0.2623 & 0.8787 & 2.4646\\
& CVAE & 0.0429 & 0.2416 & 0.5117 & 7.2728 & 11.2968 &13.1643& 0.2558& 0.8744 &2.4215\\
& EDF & \textbf{0.0597} & \textbf{0.2926} & 0.5355 & 7.9606 & 12.1776 & 13.8786 & 0.3036 & 0.8918 & 2.5842\\
& NT & \underline{0.0571} & \underline{0.2919} & 0.5424 & 8.0267 & 12.3098 & 14.0577 & \underline{0.3070} & 0.9024 & 2.6690\\
& AdaLabel & 0.0482 & 0.2573 & 0.5136 & 7.9152 & 12.0968 & 13.9496 & 0.2936 & 0.8947 & 2.6336\\
\cmidrule{2-11}
& DRESS(-NT) & 0.0554 & 0.2909 & \underline{0.5448} & \underline{8.1722} & \underline{12.5195} & \underline{14.3244} & \textbf{0.3079} & \textbf{0.9192} & \underline{2.8444}\\
& \textbf{DRESS} & 0.0547 & 0.2906 & \textbf{0.5504} & \textbf{8.1821} & \textbf{12.5533} & \textbf{14.3890} & 0.3052 & \underline{0.9153} & \textbf{2.8548}\\ 
\midrule
\midrule
\multirow{8}{*}{\shortstack[c]{Blender-90M\\(OpenSubtitles)}} &Vanilla & 0.0373 & 0.1550 & 0.2698 & 6.5882 & 9.5097 & 10.7983 & 0.1758 & 0.8459& 2.4702\\
& MMI & 0.0426 & 0.1660 & 0.2755 & 6.4854 & 9.2276 & 10.3364 & 0.2005 & 0.8721 &2.4469 \\
& CVAE & 0.0393 & 0.1804 & 0.3398 & 7.0092 & 10.5135 & 11.8959 & 0.2073 & \textbf{0.9214} & 2.5726\\
& EDF & 0.0476 & 0.2019 & 0.3536 & 7.0189 & 10.3899 & 11.8036 & 0.2161 &0.8777 &2.5738 \\
& NT & \underline{0.0504} & \underline{0.2216} & \underline{0.3969} & \underline{7.3734} & \underline{11.0928} & \underline{12.6594} & \underline{0.2480} & 0.8944 &2.7049\\
& AdaLabel & 0.0431 & 0.1913 & 0.3573 & 7.0306 & 10.5280 & 12.0680 & 0.2063 & 0.8708& 2.6407\\
\cmidrule{2-11}
& DRESS(-NT) & 0.0499 & 0.2178 & 0.3817 & 7.3316 & 10.8422 & 12.2530 & 0.2308 & 0.8927 & \underline{2.7114} \\
& \textbf{DRESS} &\textbf{0.0524} & \textbf{0.2351} & \textbf{0.4180} & \textbf{7.5113} & \textbf{11.2355} & \textbf{12.7612} & \textbf{0.2612} & \underline{0.9041} & \textbf{2.7654}\\ 
\midrule
\multirow{8}{*}{\shortstack[c]{BART-large\\(OpenSubtitles)}} & Vanilla & 0.0262 & 0.1028 & 0.1806 & 5.8507 & 8.2064 & 9.2760 & 0.1532 & 0.7803 & 2.2043\\
& MMI & 0.0275 & 0.1094 & 0.1923 & 6.0557 & 8.5303 & 9.6961 & 0.1595 & 0.8067 & 2.1626 \\
& CVAE & 0.0226 & 0.1460 & 0.3495 & 6.2232 & 9.7304 & 11.4593 & 0.1507 & 0.8600 & 2.3005\\
& EDF & \textbf{0.0474} & \underline{0.2056} & \underline{0.3572} & 7.0338 & 10.5464 & 11.9977 & 0.2209 & 0.8558 & 2.5346\\
& NT & 0.0228 & 0.0948 & 0.1594 & 5.5542 & 8.2025 & 9.6915 & 0.1165 & 0.8298 & 2.6368\\
& AdaLabel & 0.0381 & 0.1772 & 0.3316 & 7.0306 & 10.5667 & 12.0747 & 0.2030 & \underline{0.8647} & 2.5652 \\
\cmidrule{2-11}
& DRESS(-NT) & 0.0456 & 0.2006 & 0.3509 & \underline{7.1669} & \underline{10.6915} & \underline{12.1509} & \underline{0.2220} & 0.8618 & \underline{2.6620} \\
& \textbf{DRESS} & \underline{0.0472} & \textbf{0.2178 }& \textbf{0.3890} & \textbf{7.4656}& \textbf{11.2761} & \textbf{12.8601} & \textbf{0.2322} & \textbf{0.8873} & \textbf{2.7406}\\

\bottomrule
\end{tabular}

\caption{Automatic evaluation results in various diversity metrics (Dist-$n$, Ent-$n$, LF, and Sem-Ent) and coherency metric (an average MaUdE of generated responses).
\textbf{Bold} and \underline{underline} indicate the best and runner-up results, respectively.
DRESS(-NT) indicates the variant version of DRESS that only utilizes the weighted NLL without NT.
}
\vspace*{-2mm}
\label{tab:3_automatic_evaluation}
\end{table*}

\subsection{Implementation Details}\label{subsec:5_3_implementation_details}
We take two Transformer-based sequence-to-sequence models: Blender-90M~\cite{roller2021recipes} and BART-large~\cite{lewis2020bart} as the underlying generation models to demonstrate that our method widely works well on different architectures.
For DRESS, we find the optimal hyperparamters for the number of semantic clusters $k$ and the weighting factor $\gamma$ through the grid search from $k\in\{10,20,50,100\}$ and $\gamma\in\{1.0,5.0,10.0,30.0,100.0\}$, and we set them as $k$=20 and $\gamma$=30 in our whole experiments unless otherwise specified.
All models use greedy decoding strategy, and we utilize both blocking repeated $n$-grams~\cite{paulus2017deep} ($n$ = 3) within the generated response and the input sequence to prevent models from repeating subsequences.
Moreover, we release our implementation code\footnote{https://github.com/hyperconnect/sem-ent} to help researchers reproduce the result.
\section{Results and Analysis}\label{sec:6_results}
\subsection{Evaluation Results}\label{subsec:6_1_evaluation_results}
\begin{table*}[t]
\centering
\footnotesize
\begin{tabular}{lcccccc}
\toprule
\multicolumn{1}{c}{\multirow{2}[2]{*}{Comparison (A vs. B)}} & \multicolumn{3}{c}{Appropriateness} & \multicolumn{3}{c}{Informativeness} \\ \cmidrule(lr){2-4} \cmidrule(lr){5-7}
& \multicolumn{1}{c}{A wins (\%)} & \multicolumn{1}{c}{B wins (\%)}  & \multicolumn{1}{c}{Tie (\%)}  & \multicolumn{1}{c}{A wins (\%)} & \multicolumn{1}{c}{B wins (\%)}  & \multicolumn{1}{c}{Tie (\%)}  \\
\midrule
Ours vs Vanilla & \textbf{35.3} & 24.7 & 40.0 & \textbf{36.0} & 28.0 & 36.0 \\
Ours vs MMI & \textbf{40.0} & 34.7 & 25.3 & \textbf{40.7} & 36.0 & 23.3\\
Ours vs CVAE & \textbf{44.7} & 30.0 & 25.3 & \textbf{36.7} & 36.0 & 27.3 \\
Ours vs EDF & \textbf{35.3} & 24.7 & 40.0 & \textbf{32.7} & 23.3 & 44.0  \\
Ours vs NT & \textbf{28.0} & 25.3 & 46.7 & \textbf{37.3} & 26.0 & 36.7\\
Ours vs AdaLabel & \textbf{28.7} & 24.0 & 47.3 & \textbf{32.7} & 31.3 & 36.0\\
\bottomrule
\end{tabular}
\caption{Human pairwise comparison results in terms of appropriateness and informativeness of generated responses. 
The evaluation is conducted on the test set of DailyDialog with Blender-90M  using greedy decoding.}
\vspace*{-2mm}
\label{tab:4_pairwise_comparison}
\end{table*}
Table~\ref{tab:3_automatic_evaluation} shows the automatic evaluation results.
Overall, DRESS achieves the best performance in both semantic and lexical-level response diversities while showing high response coherency for most of the experimental setups.
To be more specific, DRESS shows a substantially higher semantic diversity (Sem-Ent) than all other baseline models in every experimental setup.
It is quite intriguing that DRESS also achieves better performance in lexical-level response diversity (Dist-$n$, Ent-$n$, and LF).
This observation demonstrates that improving semantic diversity entails the improvement of lexical-level diversity.
Furthermore, MaUdE results indicate that DRESS preserves better response coherency compared to other baseline methods.

Table~\ref{tab:4_pairwise_comparison} summarizes the pairwise human evaluation results.
DRESS shows better conversation abilities in appropriateness and informativeness than other baseline methods while improving response diversity (as shown in the automatic evaluation).

\subsection{Analysis of Experimental Results}\label{subsec:6_2_analysis}
\begin{figure}[t]
\centering
\includegraphics[width=\columnwidth]{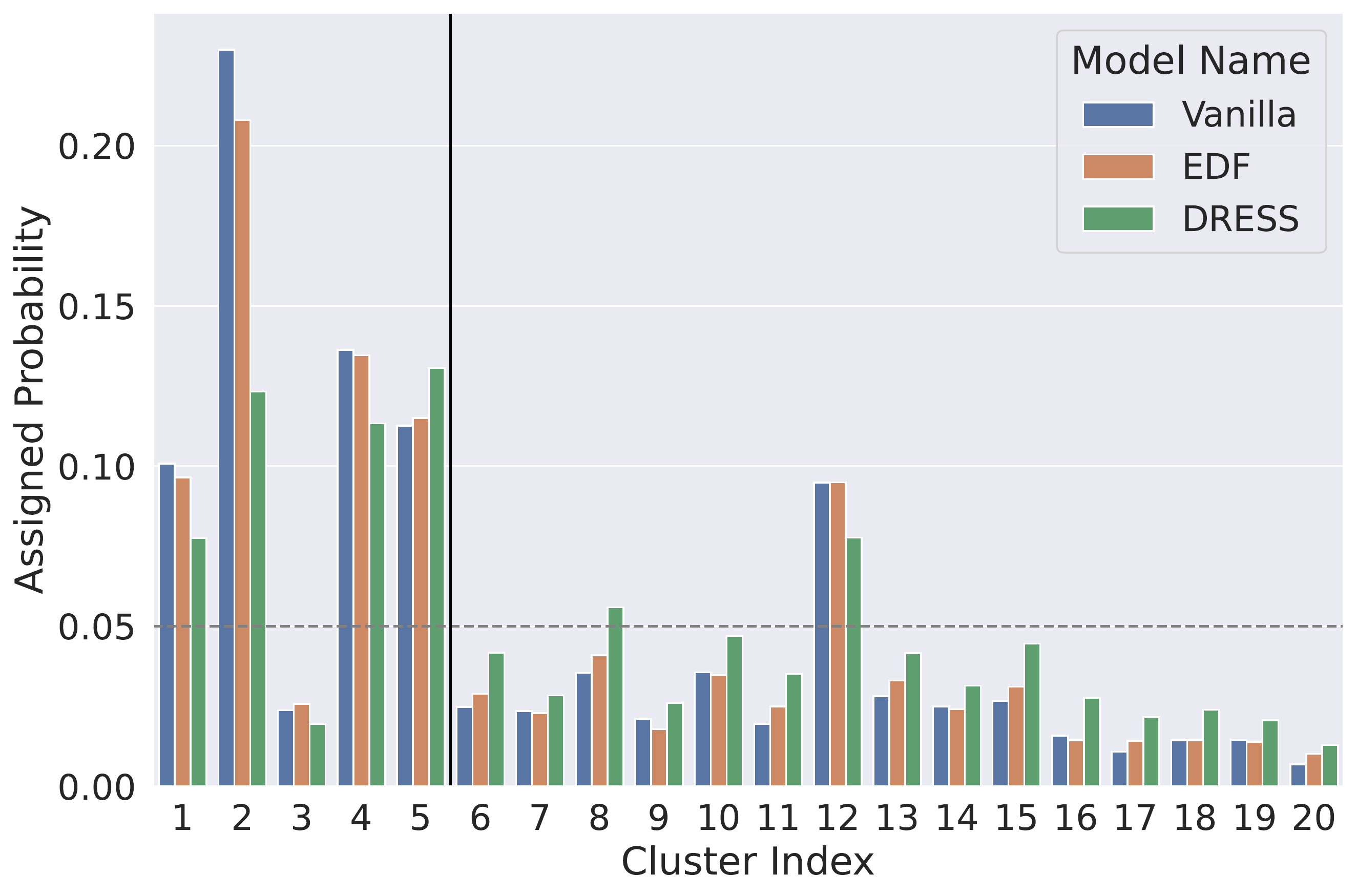}
\vspace*{-1.5em}
\caption{Probability distribution of the responses generated by Vanilla, EDF and DRESS. The dashed line indicates the uniformly distributed probability, 0.05.}
\vspace*{-1em}
\label{fig:3_prob_dist}
\end{figure}

\textbf{Semantic Distribution of Generated Responses.}
To analyze how DRESS increase the semantic diversity, we compare the semantic distribution of responses generated by DRESS with that of Vanilla and EDF.
Figure~\ref{fig:3_prob_dist} illustrates the detailed semantic distribution of the generated responses of Blender-90M on DailyDialog.
The Vanilla model shows a high probability on the head semantic clusters (e.g., Cluster 1, 2, 4) and low probability on the tail semantic clusters (e.g., Cluster 13$\sim$20).
DRESS effectively reduces the probabilities of the head semantic cluster except Cluster 5 and boosts probabilities of the tail clusters except Cluster 12, where the only one has a higher assigned probability than 0.05 in the tail clusters.

\noindent\textbf{Changing Hyperparameters of DRESS.}
As described in Section~\ref{subsec:5_3_implementation_details}, we found the optimal hyperparameter pair ($\gamma$=30, $k$=20) for Blender-90M on DailyDialog.
To examine the effect of each hyperparamter, we observe the change of evaluation results by varying the hyperparameters.
Note that we fix $\gamma$ to 30.0 and $k$ to 20 when varying $k$ and $\gamma$, respectively.
Table~\ref{tab:5_method_analysis} shows the results about the effect of the hyperparameters.
We find that increasing $\gamma$ induces dialogue generation models to produce more diverse responses, which can be shown by improvement in Dist-3, Ent-3, and Sem-Ent.
We also observe that decreasing $k$ induces dialogue generation models to generate more diverse responses.
However, MaUdE gets degraded while response diversity improves implying a trade-off between response diversity and coherence.
\newline
\textbf{Ablation Study.}
To verify the effect of our weighted NLL, we conduct an ablation study.
In Table~\ref{tab:3_automatic_evaluation}, DRESS(-NT) indicates the variant of DRESS without NT and only utilizes weighted NLL.
DRESS(-NT) shows a slight degradation in Sem-Ent compared to DRESS.
Nonetheless, DRESS(-NT) achieves better performance in Sem-Ent than other baselines excluding DRESS.
Moreover, DRESS(-NT) also shows a higher lexical-level diversity than other baselines, along with high MaUdE scores.
From these observations, we verify the effectiveness of our proposed weighted NLL that semantically diversifies generated responses.

\begin{table}[t]\setlength{\tabcolsep}{0.5em}
\centering
\footnotesize
\begin{tabular}{lcccc}
\toprule
Config & Dist-3 & Ent-3 & MaUdE & Sem-Ent \\
\midrule
$\gamma$ = 1.0 &0.4333&12.8968&0.8570 & 2.6233\\
$\gamma$ = 5.0 &0.4400 &12.9989&0.8593 &2.6551\\
$\gamma$ = 10.0 &0.4410&13.0670&0.8583 &2.6959\\
$\gamma$ = 30.0 &0.4571&13.3060&0.8575 &2.7819\\
$\gamma$ = 100.0 & 0.4625&13.5839&0.8436&2.8444\\
\midrule
$k$ = 10 & 0.4748&13.7596& 0.8390&2.8451\\
$k$ = 20 & 0.4571&13.3060&0.8575&2.7819\\
$k$ = 50 & 0.4318&13.0001& 0.8513&2.7009\\
$k$ = 100 & 0.4311&12.8857&0.8637 &2.6258\\
\bottomrule
\end{tabular}

\caption{Analyzing the effect of $\gamma$ and $k$.}
\vspace*{-2mm}
\label{tab:5_method_analysis}
\end{table}

\section{Conclusion}\label{sec:conclusion}
In this work, we present a new automatic evaluation metric, Sem-Ent, which can measure the semantic diversity of generated responses.
Sem-Ent correlates with human judgments on response diversity more than other automatic diversity metrics and shows a high correlation with human judgments on interestingness.
Moreover, we introduce a new learning method, DRESS, to mitigate the problem of the imbalanced semantic distribution of dialogue datasets.
Evaluation results show that DRESS improves both the semantic and lexical-level diversities of generated responses, along with the gain in response coherency.
\subsection*{Limitations of This Work}\label{sec:limitations}
In this section, we discuss the potential limitations of our methods and the experimental procedure.
To start with, our proposed diversity metric Sem-Ent requires a pre-trained language model to calculate the result.
This indicates that it requires relatively heavier computational resources to calculate Sem-Ent compared to other lexical-based diversity metrics such as Dist-$n$ and Ent-$n$.
Moreover, extending Sem-Ent to other languages or other domains could be problematic if no high-quality pre-trained language model is available in that language or domain.

In terms of the experimental procedure, we performed the experiment once rather than running it multiple times with different seeds.
Since our evaluation process incorporates a human annotation, which requires a payment to human annotators, we could not perform multiple sets of experiments due to the limited budget.
We could not obtain a sufficient number of annotations to acquire statistically significant results for every pairwise comparison.
In the same perspective, we conducted the human evaluation in only criteria \textit{Appropriateness} and \textit{Informativeness}.
We cannot include further criteria, such as \textit{Diversity} and \textit{Interestingness}, since these criteria require further evaluation setups requiring a considerable annotation cost.
We run an experiment only once since our evaluation requires a human evaluation which requires an extra annotation budget.
Furthermore, we only experimented with the English dialogue dataset (DailyDialog and the English portion of the OpenSubtitles).
Therefore our results do not necessarily guarantee the same result in other languages rather than English.

At last, we would like to clarify that our proposed metric, Sem-Ent, only focuses on measuring the response diversity and does not consider the response coherency.
Although this is our intention since we aim to build an unreferenced diversity metric, this limitation yields a drawback that Sem-Ent should always be jointly used with another metric that measures the response coherency (e.g., MaUdE).
Expanding Sem-Ent to consider the coherency with an input context will be an intriguing future direction for our research.
\section*{Ethical Considerations}\label{sec:ethical_consideration}
Dialogue generation models can reveal some biases and toxicities from their responses since these models leverage large-scale web-crawled data for pretraining.
This is a common consideration for works related to dialogue generation.
Moreover, while our paper focuses on diversifying responses from the semantic viewpoint, the model may unintentionally learn about offensive words while diversifying responses. 
We believe it will be meaningful to reduce potential harmful responses considering semantics in future work.

\bibliography{anthology,custom}
\bibliographystyle{acl_natbib}

\clearpage
\newpage
\appendix
\section*{Appendix}
\setcounter{section}{0}
\section{Details of Human Evaluation for Sem-Ent (Section~\ref{subsec:3_2_correlation_with_human_judgment})}\label{sec:a_details_human_evaluation_sement}
\subsection{Human Annotation}\label{subsec:a_1_human_annotation}
To collect human annotations, we use Amazon Mechanical Turk, and Figure~\ref{fig:4_human_judgment} shows the instructions and the interface for the human annotators.
To mitigate the bias from the annotators, we set a maximum number of annotations per annotator as 20 and randomly shuffle the order of the response generation settings and the corresponding response.
Since our task does not require particular expertise in linguistics, we open the annotations to non-experts.
Nonetheless, to control the annotation quality, we only allow the annotators who satisfy the following requirements: (1) HITs approval rate greater than 95\%, (2) Location is one of Australia, Canada, New Zealand, United Kingdom, and the United States, (3) Lifetime   number of HITs approved greater than 1000, as following \citet{kim2021distilling,han-etal-2022-understanding}.
We estimate that each HITs takes around 1.5 minutes on average (87 seconds per each HIT estimated by the 85th percentile of response times) and set the payment to USD 16 per hour. 
As a result, annotators are paid USD 0.40 per HITs.

\subsection{Baseline Diversity Metrics}\label{subsec:a_2_baselines}
To calculate Dist-$n$, Ent-$n$, and LF, we use \textit{NLTK} package~\cite{loper2002nltk} for tokenizing responses and preparing $n$-grams.
When calculating LF, we choose words with an occurrence count of less than 100 in each dataset.

\subsection{Further Evaluation Results on Different $n$}\label{subsec:a_4_further_results}
\begin{table*}[t]
\centering
\scriptsize
\begin{tabular}{cclllllll }
\toprule
 \textbf{Metric} & \textbf{Correlation} & \textbf{Dist-1} & \textbf{Dist-2} & \textbf{Dist-4} & \textbf{Ent-1} & \textbf{Ent-2} & \textbf{Ent-4} & \textbf{Sem-Ent}\\
\midrule 
\multirow{2}{*}{Diversity/BT} &
Pearson & -0.043 (0.918) & 0.258 (0.536) & 0.292 (0.481) & 0.518 (0.187) & 0.454 (0.257) & 0.394 (0.333) & \textbf{0.810 (0.015)} \\
& Spearman & 0.190 (0.651) & 0.380 (0.351) & 0.380 (0.351) & 0.595 (0.119) & 0.476 (0.232) & 0.380 (0.351) & \textbf{0.762 (0.028)}\\
\midrule
\multirow{2}{*}{Interesting/BT} &
Pearson & -0.143 (0.734) & 0.161 (0.702) & 0.198 (0.637) & 0.469 (0.240) & 0.392 (0.335) & 0.323 (0.434) & \textbf{0.789 (0.020)}\\
& Spearman & 0.142 (0.735) & 0.380 (0.351) & 0.380 (0.351) & 0.642 (0.085) & 0.523 (0.182) & 0.380 (0.351) & \textbf{0.667 (0.020)}\\
\bottomrule
\end{tabular}
\caption{Further evaluation results on different $n$ for Dist-$n$ and Ent-$n$.
}
\vspace{-1em}
\label{tab:6_sem_ent_human_judgment_appendix}
\end{table*}

We further report the experimental results with different $n$ for Dist-$n$ and Ent-$n$ (Table~\ref{tab:6_sem_ent_human_judgment_appendix}).
Like the results in Table~\ref{tab:1_sem_ent_human_judgment}, our proposed Sem-Ent also shows the higher correlation with human judgements on response diversity than Dist-$n$ and Ent-$n$.

\subsection{Why We Did Not Compare Sem-Ent with NLI Diversity~\citep{stasaski2022semantic}}\label{subsec:a_5_why}
\begin{figure*}[t]
\centering
\includegraphics[width=0.95\textwidth]{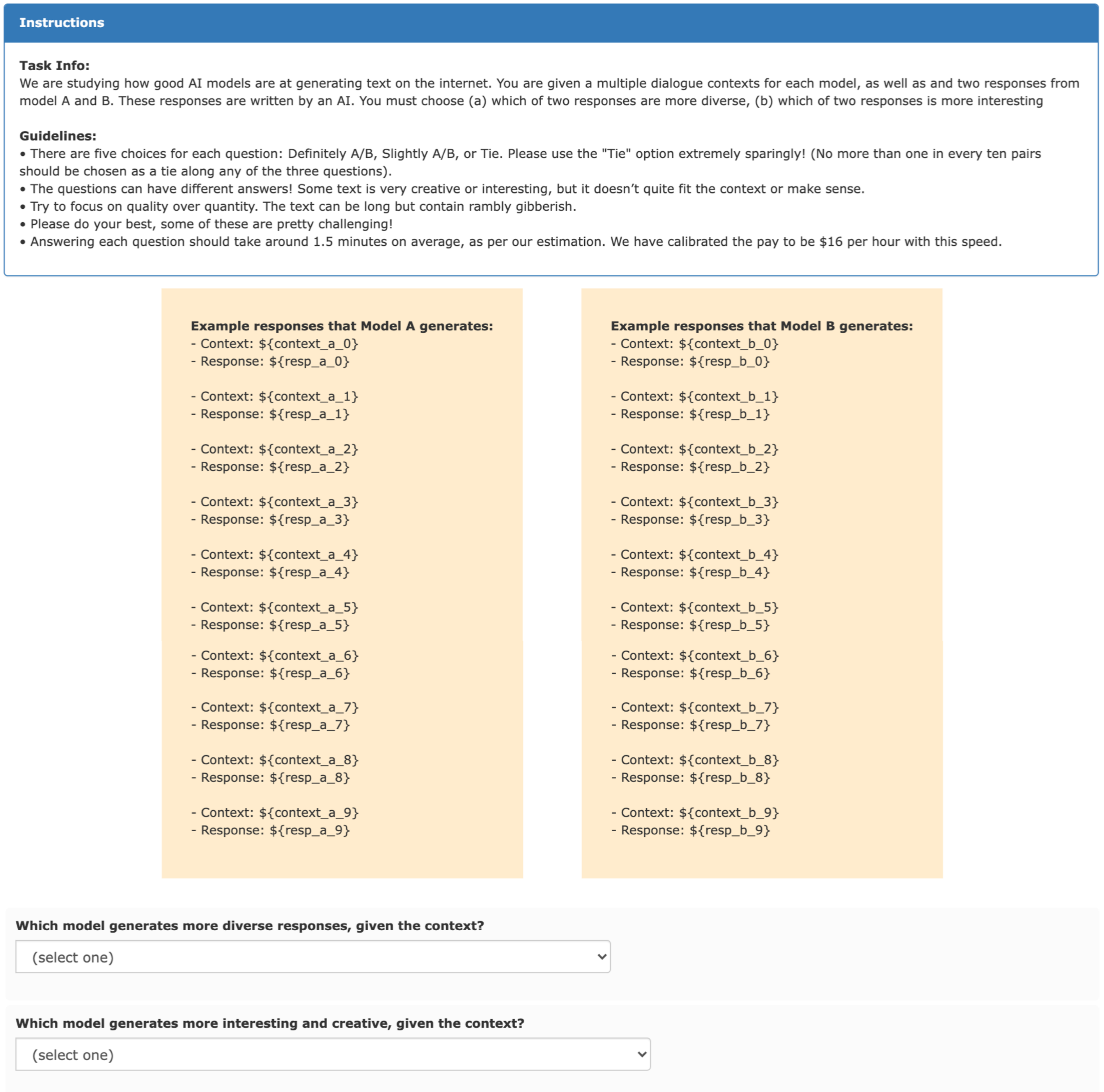}
\caption{The interface of human evaluation for assessing how responses are (a) diverse, (b) more interesting and creative.}
\label{fig:4_human_judgment}
\vspace*{-2mm}
\end{figure*}

As we propose a new diversity metric (Sem-Ent), we also understand that it is required to compare Sem-Ent with a recently proposed diversity metric (NLI Diversity) to verify its effectiveness.
However, we would like to emphasize that the two diversity metrics target the different evaluation settings.
NLI Diversity targets the sample-wise semantic diversity setting where the metric measures whether semantically diverse responses are generated for a single test sample.
Thus, \citet{stasaski2022semantic} conducted experiments on the McDiv benchmark~\cite{tevet2021evaluating}, which evaluates the semantic diversity metrics in the sample-wise semantic diversity setting to show the effectiveness of NLI Diversity.
On the other hand, our Sem-Ent targets the overall semantic diversity setting that assesses the diversity of generated responses throughout the test set.
A higher Sem-Ent score indicates that the generation model covers a wide range of semantic topics throughout the test set.
Therefore, our human evaluation simulates the overall semantic diversity setting, unlike the McDiv benchmark, as shown in Figure~\ref{fig:4_human_judgment}.
Note that NLI Diversity is incompatible with this evaluation setting because dialogue generation models should produce multiple responses for each test sample.
Moreover, there is no guarantee that generated responses throughout the test set will be semantically diverse even if dialogue generation models show high diversity for each test sample.
We believe that our proposed Sem-Ent and NLI Diversity can be used complementary to each other in evaluating the response diversity without conflict.

\section{Robustness of Sem-Ent (Section~\ref{subsec:3_2_correlation_with_human_judgment})}\label{sec:b_robustness}
We examine the robustness of Sem-Ent by changing the configurations used for calculating the metric.
Several configurations can be changed in Sem-Ent, including the types of language models for mapping responses $r$ into a semantic representation $e(r)$ and the number of clusters $k$ for the $k$-means algorithm.
Varying the configurations, we compute Sem-Ent on responses generated by Blender-90M~\citep{roller2021recipes} for the test set of DailyDialog~\citep{li2017dailydialog} with all methods (in Table~\ref{tab:3_automatic_evaluation}).
We then measure the Spearman rank correlation between the computed Sem-Ent scores of the different configurations.

For the choice of language models, we compare three variants: DialoGPT~\citep{zhang2020dialogpt}, RoBERTa \citep{liu2019roberta}, and GPT2-large \citep{radford2019language}.
The average Spearman rank correlation between the pairs of these three variants (3 pairs) is 0.8809.
For the number of clusters, we vary the number $k$ with values in $\{10, 20, 50, 100\}$ and compare the scores ranked by Sem-Ent.
The average Spearman rank correlation between these configurations (6 pairs) is 0.9821.
High correlations show that Sem-Ent produces similar rankings of different models regardless of different configurations, indicating that Sem-Ent is a robust diversity metric against the choice of configurations.

\section{Details of Human Evaluation for DRESS (Section~\ref{sec:5_experiments_dress})}\label{sec:c_details_human_evaluation_dress}
\begin{figure}[t]
\centering
\includegraphics[width=\columnwidth]{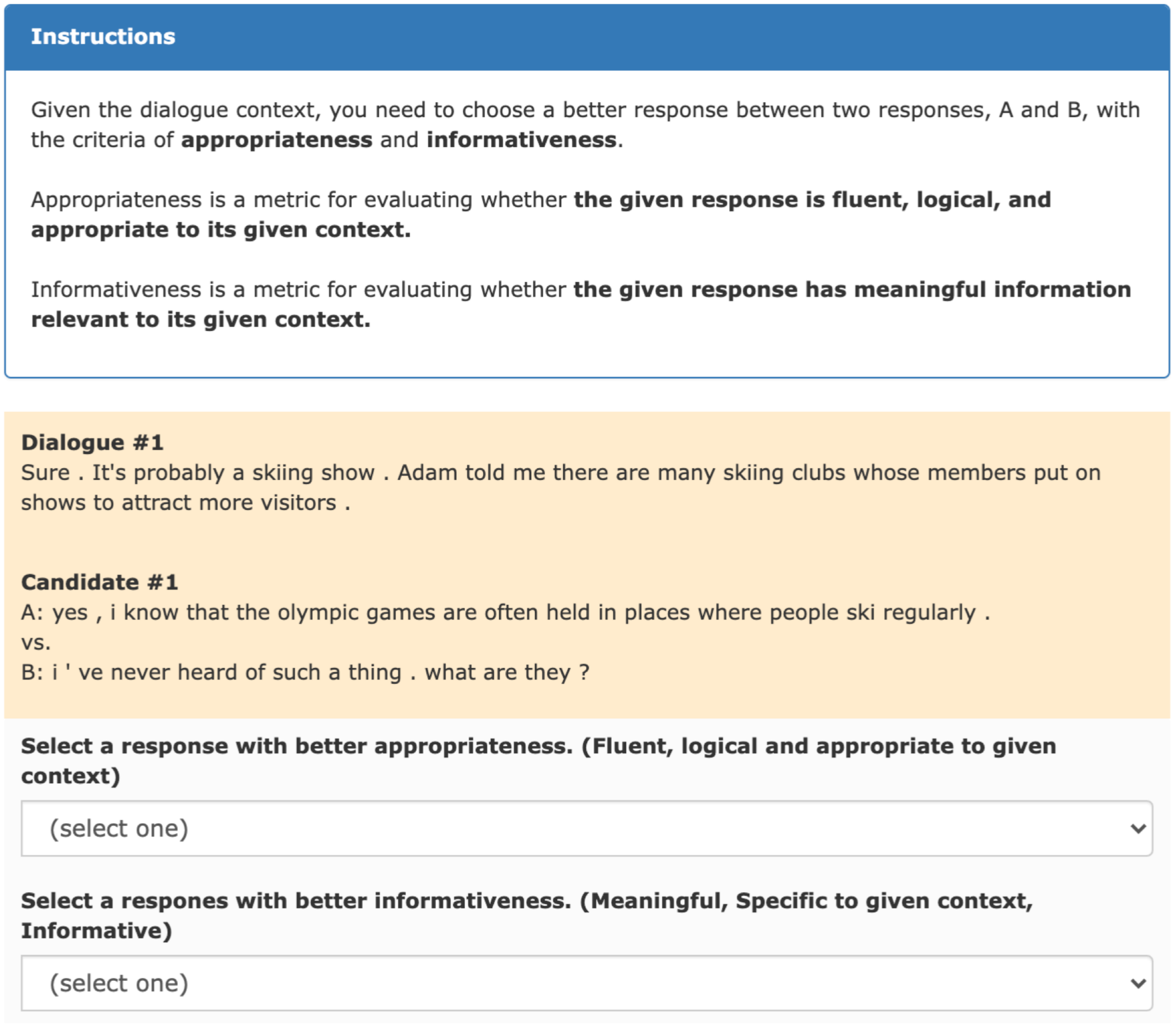}
\caption{The interface of pairwise human evaluation for appropriateness and informativeness.}
\label{fig:5_human_pairwise}
\end{figure}
\subsection{Human Annotation}\label{subsec:c_1_human_annotation}
To collect human annotations for verifying the effectiveness of our proposed DRESS, we also use Amazon Mechanical Turk and the same setting in Section~\ref{subsec:a_1_human_annotation} to mitigate the bias and control the quality of the human annotations.
Figure~\ref{fig:5_human_pairwise} shows the instructions and the interface for the human evaluation. 
Here, human annotators are paid USD 0.25 per HITs as we estimate that each HITs takes around 1.4 minutes on average (84 seconds per HITs estimated by the 85th percentile of response times) and set the payment to USD 10.7 per hour since the difficulty of the this annotation is easier than the human annotations for Sem-Eet in Section~\ref{subsec:a_1_human_annotation}.

\subsection{Why We Could Not Set \textit{Diversity} as a Human Evaluation Criterion}\label{subsec:c_2_criterion}
Unlike \textit{Appropriateness} and \textit{Informativeness}, annotators require multiple conversations to annotate \textit{Diversity} of generation models because it is hard to capture the response diversity from a single conversation.
Additional Amazon Mechanical Turk sessions should be conducted besides our pairwise human evaluation to obtain annotations on \textit{Diversity}.
However, as described in Limitation Section, it was challenging under our limited human evaluation budget.
Therefore, we conducted a pairwise human evaluation on only \textit{Appropriateness} and \textit{Informativeness} following the most common evaluation setup, as shown in Figure~\ref{fig:5_human_pairwise}.
We hope that this experimental result will be used as a reference to help understand the general conversational ability of our proposed DRESS, even though DRESS is proposed to increase semantic diversity.

\section{Implementation Details (Section~\ref{subsec:3_2_correlation_with_human_judgment} and Section~\ref{sec:5_experiments_dress})}\label{sec:d_implementation_details}
\subsection{Training Models}\label{subsec:d_1_models}
All of our experiments are done using the ParlAI~\cite{miller2017parlai} framework.
We leverage model weights of Blender-90M and BART-large provided in the ParlAI framework.
Blender-90M is pretrained on Reddit corpus, and BART-large is pretrained jointly on Wikipedia and Toronto Books.
Note that Blender-90M has 90M parameters and BART-large consists of 400M parameters.
All baselines and DRESS use the initial learning rate of $7e$-$6$ with Adam optimizer, except CVAE for Blender-90M trained on DailyDialog using $2e$-$5$, MMI for Blender-90M trained on OpenSubtitles using $1e$-$6$, and CVAE for Blender-90M trained on OpenSubtitles using $1e$-$5$.
We search the appropriate learning rate for those exceptions since those exceptions are not stable enough to train the model.
We use a learning rate scheduler that reduces its learning rate by multiplying 0.5 when the loss has stopped decreasing.
All Blender-90M models and all BART-large models are trained using a batch size of 32 and 16 on a single A100 GPU, respectively.
Training a single model takes less than a day with these configurations.
\subsection{Language Models for Calculating Sem-Ent}\label{subsec:d_2_language_models}
In this work, we test three language models to obtain semantic representations of responses: DialoGPT, RoBERTa, and GPT2-large.
For reproducibility, we utilize model weights, which are publicly provided in HuggingFace Transformers~\cite{wolf2020transformers}: \texttt{microsoft/DialoGPT-large}, \texttt{roberta-base}, and \texttt{gpt2-large}, for DialoGPT, RoBERTa, and GPT2-large, respectively.
\subsection{Software and Hardware}\label{subsec:d_3_environment}
We use Python 3.8, PyTorch 1.9.0 (py3.8\_cuda11.1\_cudnn8.0.5\_0), HuggingFace Transformers 4.6.1, and ParlAI 1.3.0.
All the experiments are done using NVIDIA A100-40GB GPUs, along with AMD EPYC 7742 64-Core Processors.
\subsection{License}\label{subsec:d_4_licence}
The DailyDialog dataset has CC-BY-NC-SA 4.0 license.
OpenSubtitles dataset does not specify the license on the dataset.
For the pretrained models, DialoGPT, RoBERTa, and GPT-2 large are all released with the MIT license.
Since CC-BY-NC-SA 4.0 and MIT license both allow the resource utilization for research purposes, the use of these scientific artifacts in this work is valid. 

\section{Further Analysis (Section~\ref{sec:6_results})}\label{sec:f_further_analysis}
\subsection{Confidence Interval of MaUdE Scores}\label{subsec:e_1_maude}
\begin{table}[t]\setlength{\tabcolsep}{0.4em}
\centering
\footnotesize
\begin{tabular}{llc}
\toprule
Backbone& Method & MaUdE ($\pm$ 95\% CI)\\
\midrule
\multirow{8}{*}{\shortstack[c]{Blender-90M\\(DailyDialog)}} & Vanilla & 0.8489 $\pm$ 0.005\\
& MMI & 0.8208 $\pm$ 0.005\\
& CVAE & 0.8552 $\pm$ 0.005\\
& EDF & \underline{0.8593} $\pm$ 0.005\\
& NT & \textbf{0.8597} $\pm$ 0.005\\
& AdaLabel & 0.8443 $\pm$ 0.005\\
\cmidrule{2-3}
& DRESS(-NT) & 0.8460 $\pm$ 0.005\\
& \textbf{DRESS} & 0.8575 $\pm$ 0.005\\ 
\midrule
\multirow{8}{*}{\shortstack[c]{BART-large\\(DailyDialog)}} & Vanilla & 0.8854 $\pm$ 0.005\\
& MMI & 0.8787 $\pm$ 0.005\\
& CVAE& 0.8744 $\pm$ 0.005\\
& EDF & 0.8918 $\pm$ 0.004\\
& NT & 0.9024 $\pm$ 0.004\\
& AdaLabel& 0.8947 $\pm$ 0.004\\
\cmidrule{2-3}
& DRESS(-NT) & \textbf{0.9192} $\pm$ 0.003\\
& \textbf{DRESS} & \underline{0.9153} $\pm$ 0.003\\ 
\midrule
\midrule
\multirow{8}{*}{\shortstack[c]{Blender-90M\\(OpenSubtitles)}} &Vanilla & 0.8459  $\pm$ 0.004\\
& MMI & 0.8721 $\pm$ 0.004\\
& CVAE  & \textbf{0.9214} $\pm$ 0.003 \\
& EDF & 0.8777 $\pm$ 0.004\\
& NT  & 0.8944  $\pm$ 0.003\\
& AdaLabel  & 0.8708  $\pm$ 0.004\\
\cmidrule{2-3}
& DRESS(-NT) & 0.8927 $\pm$ 0.003\\
& \textbf{DRESS} & \underline{0.9041} $\pm$ 0.003\\ 
\midrule
\multirow{8}{*}{\shortstack[c]{BART-large\\(OpenSubtitles)}} & Vanilla & 0.7803  $\pm$ 0.005\\
& MMI & 0.8067 $\pm$ 0.005\\
& CVAE & 0.8600  $\pm$ 0.004\\
& EDF & 0.8558 $\pm$ 0.004\\
& NT & 0.8298  $\pm$ 0.005\\
& AdaLabel & \underline{0.8647}  $\pm$ 0.004\\
\cmidrule{2-3}
& DRESS(-NT)  & 0.8618 $\pm$ 0.004\\
& \textbf{DRESS} & \textbf{0.8873} $\pm$ 0.003\\

\bottomrule
\end{tabular}

\caption{MaUdE with a 95\% confidence interval when automatically evaluating various methods.}
\vspace*{-2mm}
\label{tab:7_automatic_maude_ci}
\end{table}

\begin{table}[t]
\setlength{\tabcolsep}{0.5em}
\centering
\footnotesize
\begin{tabular}{lc}
\toprule
Config & MaUdE ($\pm$ 95\% CI) \\
\midrule
$\gamma$ = 1.0 & 0.8570 $\pm$ 0.004 \\
$\gamma$ = 5.0 & 0.8593 $\pm$ 0.004 \\
$\gamma$ = 10.0 &0.8583 $\pm$ 0.004\\
$\gamma$ = 30.0 &0.8575 $\pm$ 0.004\\
$\gamma$ = 100.0 &0.8436 $\pm$ 0.004\\
\midrule
$k$ = 10 & 0.8390 $\pm$ 0.004\\
$k$ = 20 & 0.8575 $\pm$ 0.004\\
$k$ = 50 & 0.8513 $\pm$ 0.004\\
$k$ = 100 & 0.8637 $\pm$ 0.004\\
\bottomrule
\end{tabular}

\caption{MaUdE with a 95\% confidence interval when analysing the effect of hyperparameters, $\gamma$ and $k$.}
\vspace*{-2mm}
\label{tab:8_analysis_ci}
\end{table}

In Table~\ref{tab:3_automatic_evaluation} and Table~\ref{tab:5_method_analysis}, we report the average MaUdE score of responses generated by each method.
To provide descriptive statistics of evaluation, here we provide a 95\% confidence interval of MaUdE in Table~\ref{tab:7_automatic_maude_ci} and Table~\ref{tab:8_analysis_ci}.
Note that we only report confidence intervals of MaUdE since other diversity metrics (Dist-$n$, Ent-$n$, LF, Sem-Ent) return a single value from a set of responses.
Thus, we can not calculate the confidence interval of the diversity metrics.

\subsection{Inter-Annotator Reliability of Pairwise Human Evaluation}\label{subsec:e_2_reliability}
We calculate a Fleiss' Kappa for pairwise human evaluation results to measure the annotation variance.
We find that Fleiss' Kappas are 0.09 and 0.04 for appropriateness and informativeness, respectively.
Although these values are not high, as~\citet{kulikov2019importance} and~\citet{wong2021cross} show that inter-annotator reliability of annotation results using crowd-sourced annotators (such as in our case, using Amazon Mechanical Turk) can be low since annotators show high cultural and training variances, especially when the task is subjective as our case.
Note that 64 annotators participated in our human evaluation, and we limited the number of maximum annotations that a single annotator can be assigned to reduce the bias, which might have increased inter-annotator diversity.

\subsection{Additional Response Examples in Figure~\ref{fig:3_prob_dist} (Section~\ref{subsec:6_2_analysis})}\label{subsec:e_3_additional_example}
\begin{table}[t]
\centering
\footnotesize
\begin{tabular}{c|p{0.78\columnwidth}}
\toprule
\textbf{Index} & \textbf{Responses} \\
\midrule 
\multirow{5}{*}{3} & 
 $\boldsymbol{\cdot}$ That is wonderful .\\
& $\boldsymbol{\cdot}$ That is a wonderful choice .  \\
& $\boldsymbol{\cdot}$ Sounds painful ! \\
& $\boldsymbol{\cdot}$ Thanks . That ' s terrific ! \\
& $\boldsymbol{\cdot}$ Thanks . This is fun ! \\

\midrule
\multirow{5}{*}{5} & 
 $\boldsymbol{\cdot}$ OK, It's here , one of the best makes in China .\\
& $\boldsymbol{\cdot}$ Well , my whole family is in the United States now , but we're from Costa Rica originally .\\
& $\boldsymbol{\cdot}$ I surely do . They must have had advanced machines in ancient China to do that . \\
\midrule
\multirow{7}{*}{12} &
$\boldsymbol{\cdot}$ Here is your change and your receipt . Do you have goods unpaid on you , sir . \\
& $\boldsymbol{\cdot}$ We found your samples very attractive . We ' re interested in buying your garments if your prices are reasonable . \\
& $\boldsymbol{\cdot}$ A postcard costs you five yuan . A dozen postcards cost you 60 yuan . \\
\bottomrule
\end{tabular}
\caption{Additional response examples of the semantic clusters of the test set in the DailyDialog dataset. \textit{Index} column indicates the Cluster Index in Figure \ref{fig:2_dataset}.}
\vspace*{-1em}
\label{tab:9_additional_examples}
\end{table}

We observe the unexpected evaluation results in Figure~\ref{fig:3_prob_dist}.
Cluster 3 shows a lower assigned probability than 0.05. 
On the other hand, Cluster 12 shows a higher assigned probability than 0.05.
Unexpectedly, DRESS increases the assigned probability of Cluster 5 in the head clusters.
Therefore, we provide response examples in the corresponding clusters, in Table~\ref{tab:9_additional_examples}.
Cluster 3 includes the many generic responses still and Clusters 5 \& 12 include the responses in specific topics such as countries (Cluster 5) and purchase (Cluster 12).
Unfortunately, we fail to find the reasons for the unexpected experimental results although we examine the generated responses.
We conjecture that there is a gap between the semantic distributions of contexts in the training set and the test set in the DailyDialog dataset where the test set includes many contexts belonging to Cluster 5 \& 12 than the training set.

\subsection{Analysis of the Distribution of Generated Responses}
\begin{figure}[t]
\centering
\includegraphics[width=\columnwidth]{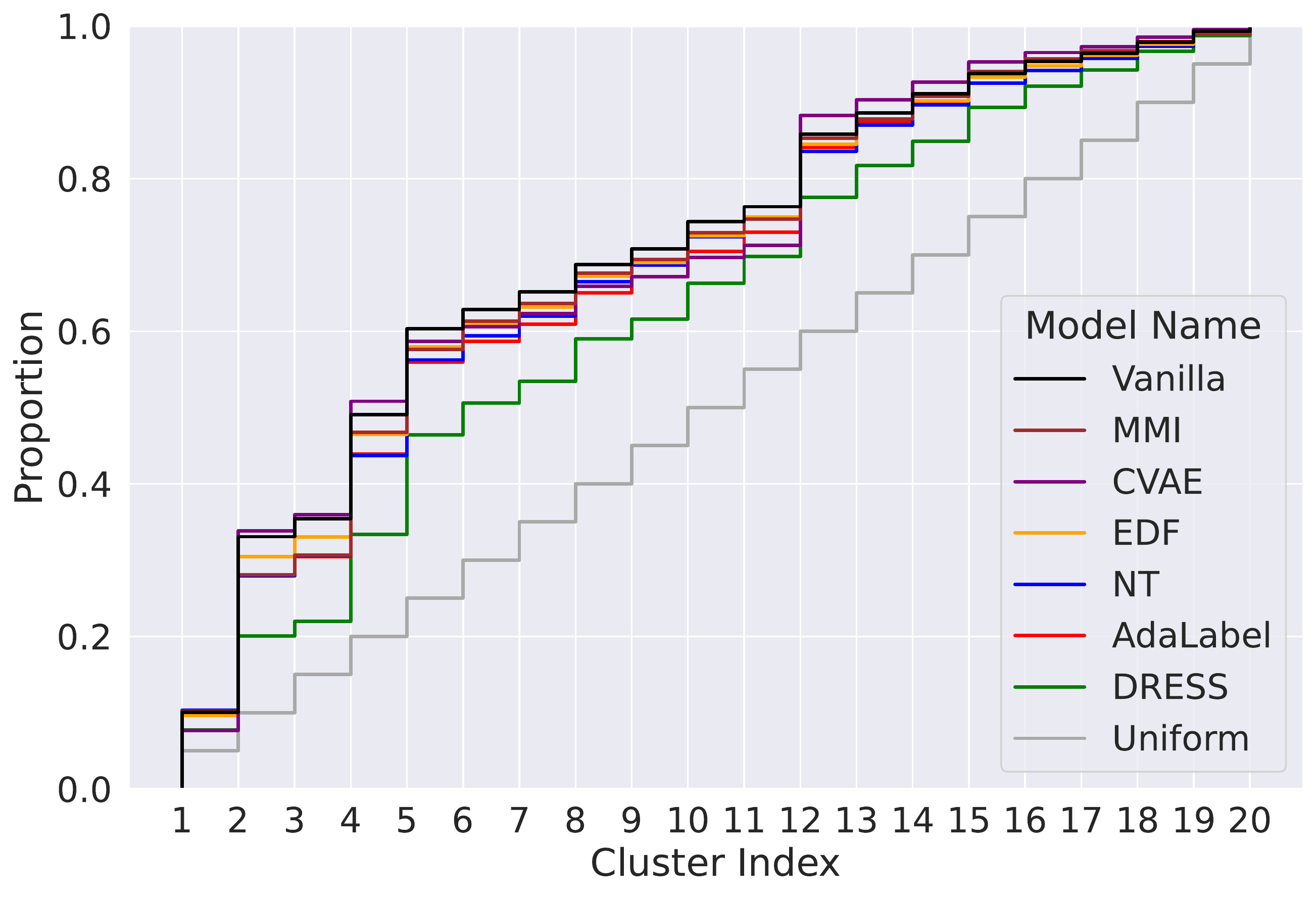}
\caption{Cumulative probability distribution of the responses generated by different methods. \textit{Uniform} illustrates the case of uniform cluster distribution.}
\label{fig:6_cum_prob_dist}
\end{figure}
Figure~\ref{fig:6_cum_prob_dist} illustrates the cumulative semantic probability distributions of the generated responses.
DRESS clearly shows the most similar cumulative distribution to that of uniform distribution, which is a distribution that achieves the highest Sem-Ent value.
Moreover, DRESS dramatically reduces the distribution of the head clusters containing generic responses compared to other baseline methods and conversely enlarges the distribution of the tail clusters.

\end{document}